%% file: iclr2025_conference.tex
\documentclass{article} %

\usepackage[preprint]{neurips_2025}
\usepackage{times}
\usepackage{booktabs}
\usepackage{graphicx}
\usepackage{xspace}
\usepackage{color,xcolor}
\usepackage{makecell}
\usepackage{colortbl}
\usepackage{multirow}
\input{math_commands.tex}

\usepackage{hyperref}
\usepackage{url}

\newcommand{\ourdata}{\textsc{GPT-Image-Edit-1.5M}\xspace}

\title{\ourdata \\ A \textit{Million-Scale}, \textit{GPT-Generated} Image Dataset}

\author{%
  Yuhan Wang$^1$ \, \, Siwei Yang$^1$ \, \, Bingchen Zhao$^2$ \, \, Letian Zhang$^1$ \vspace{.1em}\\ \textbf{Qing Liu$^3$} \, \, \textbf{Yuyin Zhou$^1$} \, \, 
  \textbf{Cihang Xie$^1$} \vspace{.5em}\\ 
 $^1$University of California, Santa Cruz \,\,\,\,\, $^2$The University of Edinburgh \,\,\,\,\, $^3$Adobe \vspace{.5em}
  \\
  \small
 \includegraphics[height=1.1em]{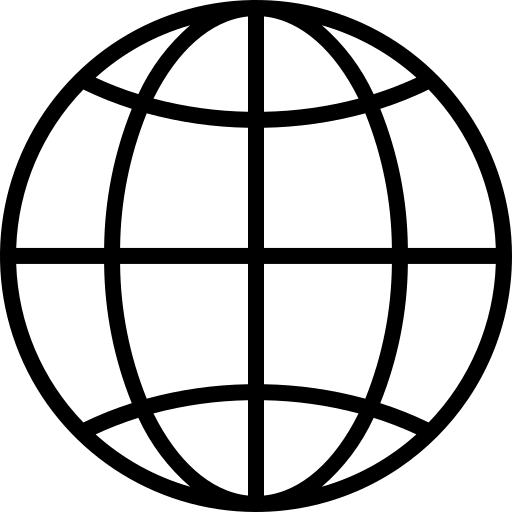} \textbf{Project Page}: \url{https://ucsc-vlaa.github.io/GPT-Image-Edit} \\
  \small
  \includegraphics[height=1.2em]{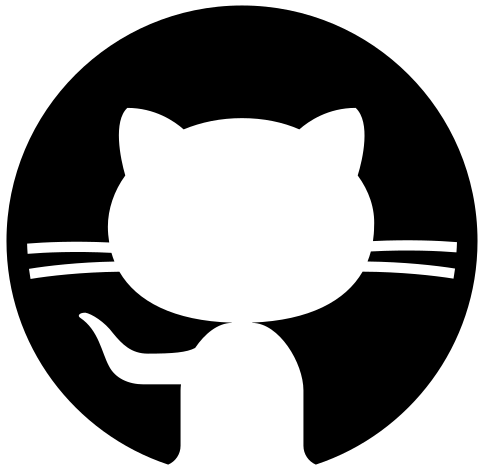} \textbf{Code}: \url{https://github.com/wyhlovecpp/GPT-Image-Edit} \\
  \small
  \includegraphics[height=1.2em]{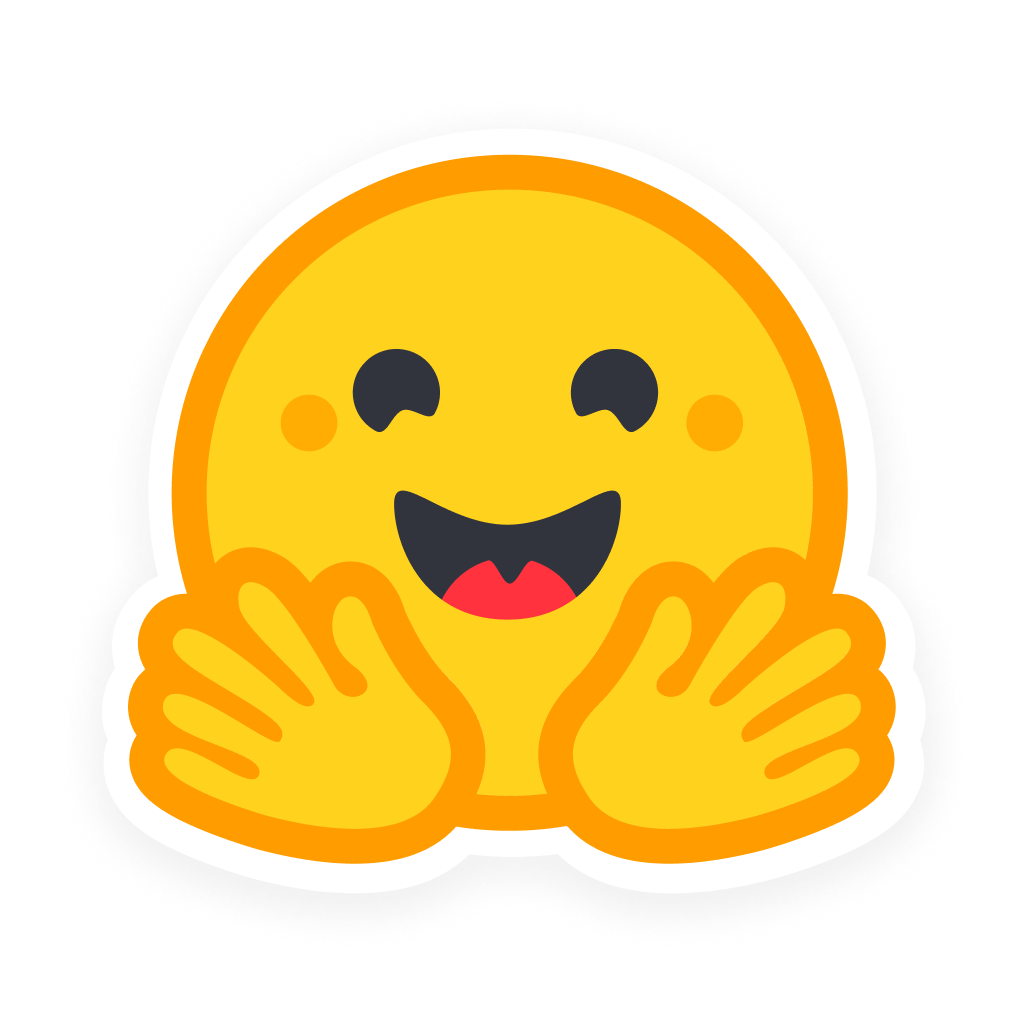} \textbf{Dataset}: \url{https://huggingface.co/datasets/UCSC-VLAA/GPT-Image-Edit-1.5M}
  \vspace{-1.6em}
}

\definecolor{mygray}{gray}{.9}

\begin{document}

\maketitle
\thispagestyle{plain} \pagestyle{plain}
\begin{abstract}
Recent advancements in large multimodal models like GPT-4o have set a new standard for high-fidelity, instruction-guided image editing. 
However, the proprietary nature of these models and their training data creates a significant barrier for open-source research. 
To bridge this gap, we introduce \ourdata, a publicly available, large-scale image-editing corpus containing more than 1.5 million high-quality triplets \textit{\{instruction, source image, edited image\}}. 
We systematically construct this dataset by leveraging the versatile capabilities of GPT-4o to \textit{unify} and \textit{refine} three popular image-editing datasets: OmniEdit, HQ-Edit, and UltraEdit. 
Specifically, our methodology involves 1) regenerating output images to enhance visual quality and instruction alignment, and 2) selectively rewriting prompts to improve semantic clarity. 
To validate the efficacy of our dataset, we fine-tune advanced open-source models on \ourdata. 
The empirical results are exciting --- \emph{e.g.}, the  \textbf{fine-tuned FluxKontext} achieves \textit{highly competitive} performance across a comprehensive suite of benchmarks, including \textbf{7.24}@GEdit-EN, \textbf{3.80}@ImgEdit-Full, and \textbf{8.78}@Complex-Edit, showing stronger instruction following and higher perceptual quality while maintaining identity.  These scores markedly exceed all previously published open-source methods and substantially narrow the gap to leading proprietary models. 
We hope the full release of \ourdata can help to catalyze further open research in instruction-guided image editing.

\end{abstract}

\section{Introduction}

Instruction-guided image editing has rapidly emerged as a key research direction in generative AI, 
with a series of diffusion- and inversion-based methods demonstrating that natural-language instructions can drive high-quality, controllable edits 
(\emph{e.g.}, InstructPix2Pix \citep{brooks2023instructpix2pix}, Prompt-to-Prompt \citep{hertz2022prompt}, SDEdit \citep{meng2021sdedit}, Imagic \citep{kawar2023imagic}). Among all models,
proprietary systems, exemplified by GPT-4o \citep{hurst2024gpt}, have established a particularly high benchmark for performance, executing edits with impressive semantic understanding and photorealism.
However, the closed-source nature of these leading models \citep{shi2024seededit,wang2025seededit} and the datasets they are trained on limits the progress of the broader research community, creating a disparity between proprietary capabilities and open-source alternatives. 

A primary bottleneck for developing powerful open-source image editing models is the absence of sufficiently large, diverse, and high-quality training data. 
Existing public resources (\emph{e.g.}, OmniEdit~\citep{wei2025omniedit}, HQ-Edit~\citep{hui2025hqedit}, UltraEdit~\citep{zhao2024ultraedit}) sometimes contain noisy or simplistic instructions, misaligned input–output pairs, or limited editing diversity between the user's intent and the resulting edited image. 
This data gap makes it challenging to train models that can rival the nuanced understanding and execution quality of their closed-source counterparts. \citep{wei2025omniedit,zhao2024ultraedit,lin2025uniworld}

\begin{figure}[t!]
\begin{center}
\includegraphics[width=.95\linewidth]{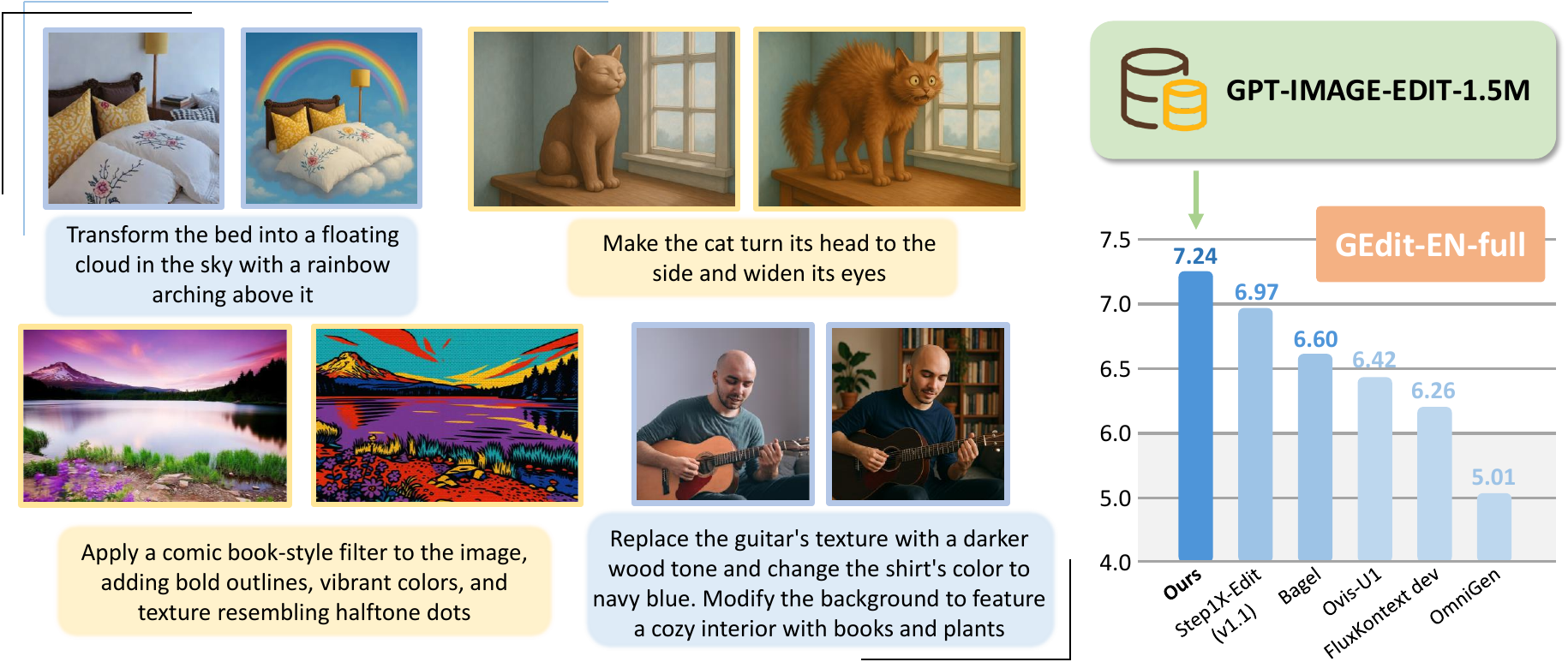}
\end{center}
\vspace{-1em}
\caption{An overview of the \ourdata dataset. The figure presents qualitative examples from our dataset, showcasing its ability to handle complex and diverse instruction-guided edits. The bar chart on the right demonstrates the effectiveness of our data; a model fine-tuned on \ourdata achieves a new state-of-the-art score of 7.24 on the GEdit-EN-full benchmark, outperforming existing open-source methods.}
\label{fig:teaser}
\vspace{-1em}
\end{figure}

To address this challenge, we introduce \ourdata, a large-scale dataset of over 1.5 million samples designed to propel open-source image editing forward. 
An overview of \ourdata is presented in Fig~\ref{fig:teaser}.
Our core strategy is to leverage a frontier model, GPT-4o, as a data generation and refinement tool. 
We unify and enhance three prominent datasets --- OmniEdit \citep{wei2025omniedit}, HQ-Edit \citep{hui2025hqedit}, and UltraEdit \citep{zhao2024ultraedit} --- through a systematic, multi-faceted approach. 
Specifically, our initial step involved \emph{regenerating only the output images} for existing pairs to improve visual quality and alignment. 
This alone provided a significant boost; for example, on the OmniEdit dataset \citep{wei2025omniedit}, fine-tuning Flux 1.0 dev \citep{flux2024} on this regenerated data (`omniedit100k-gpt') improved the imgedit score from 2.94 to 3.24 compared to the baseline. 

Nonetheless, we observed that GPT-4o would sometimes interpret instructions creatively, causing a subtle semantic drift between the original instruction and the new output.
To correct this instruction-image mismatch, we further developed two more sophisticated data refinement techniques. 
First, we introduced \emph{instruction regeneration}, where we prompted GPT-4o to write a new, more accurate instruction describing the transformation from the original input to the newly generated output. 
This `gpt-rewrite' variant further improved performance, raising the imgedit \citep{ye2025imgedit} score on OmniEdit to 3.40. 
Second, inspired by the HQ-Edit \citep{hui2025hqedit} methodology, we implemented \emph{full pair regeneration}, where both the input and output images are synthesized by the model. 
This `pair-regen' strategy, applied to the HQ-Edit subset \citep{hui2025hqedit}, also demonstrated consistent gains, lifting the GEdit-EN \citep{liu2025step1x} score from 5.67 to 5.73 on the Flux model. 
This iterative refinement process—from output regeneration to instruction rewriting and full pair synthesis—was crucial for creating a dataset with high-fidelity alignment between text and images.

Lastly, we systematically demonstrate the profound impact of our \ourdata dataset by fine-tuning a powerful open-source model, FluxKontext dev \citep{labs2025flux}. 
By additionally enhancing its architecture with Qwen-VL-7b \citep{bai2025qwen2} embeddings for improved condition-image alignment, our model achieves exciting performance on a wide array of benchmarks. 
As shown in our experiments, training on \ourdata yields substantial performance gains over models trained on the original datasets, with our final model achieving an average score of 7.236 on GEdit-EN \citep{liu2025step1x} and 3.80 on ImgEdit-Full \citep{ye2025imgedit}, placing it among the top-performing open-source models. 
Our ablation studies confirm that data regenerated by GPT-4o is the key driver of this improvement. 
Interestingly, our findings also suggest that simply increasing instruction complexity without ensuring identity preservation can be counterproductive, highlighting the critical importance of data quality in the training process. 
By making this \ourdata dataset and the trained model publicly available, we hope to provide a valuable resource for the community to keep pushing the frontier of open research in image editing.

\section{Related Works}
\label{sec:related_works}

\paragraph{Instruction-Guided Image Editing.}
The paradigm of instruction-guided image editing was largely established by InstructPix2Pix \citep{brooks2023instructpix2pix}, which framed the task as a supervised learning problem.  This seminal work introduced a scalable, two-stage pipeline: first, generating a large synthetic dataset of \textit{\{instruction, source image, edited image\}} triplets by prompting a large language model (GPT-3) to create editing instructions~\citep{brown2020language}, and then using a text-to-image model (Stable Diffusion) with Prompt-to-Prompt controls to generate the corresponding image pairs~\citep{hertz2022prompt}.  While foundational, the performance of InstructPix2Pix was inherently capped by the generative capabilities of its underlying models --- latent-diffusion-based architectures trained with CLIP --- which struggled with photorealism and complex semantic understanding~\citep{rombach2022high}.  This limitation spurred subsequent research to focus on two primary vectors of improvement: the quality and complexity of training data, and the power of the underlying generative architecture.

\paragraph{Data-Centric Advancements.}
Recognizing that data quality is a critical bottleneck, recent efforts have shifted towards more sophisticated data curation strategies.  Works like HQ-Edit~\citep{hui2025hqedit} leveraged more powerful proprietary models, namely GPT-4V \citep{hurst2024gpt} and DALL-E~3 \citep{OpenAI2023DALL-E-3}, to generate higher-fidelity and better-aligned image–instruction pairs from scratch.  
A parallel and highly effective trend is the direct distillation of capabilities from frontier models.  ShareGPT-4o-Image~\citep{chen2025sharegpt} exemplifies this by using GPT-4o's own image generation API to create a high-quality dataset of over 90,000 text-to-image and image-editing samples, with the explicit goal of transferring its advanced skills to smaller, open-source models.  Our work aligns with this data-centric philosophy, utilizing GPT-4o not just for generation, but for the systematic refinement and enhancement of existing large-scale datasets.

\paragraph{Architectural Evolution: From Diffusion to Flow Matching.}
The backbone of generative models has undergone significant evolution.  Early methods were built upon U-Net-based diffusion models~\citep{rombach2022high}, which, while effective, have limitations in modeling long-range dependencies.  A major architectural shift came with Diffusion Transformers (DiT)~\citep{peebles2023scalable}, which replaced the U-Net with a more scalable transformer architecture operating on latent patches and demonstrated superior performance as model size and compute increase.  More recently, flow matching (FM) models have emerged as a more efficient alternative to traditional diffusion~\citep{lipman2022flow}.  Instead of learning a multi-step denoising process, FM models learn to predict a continuous velocity field that transforms a simple prior distribution directly into the target data distribution by solving an ordinary differential equation.  Our choice to fine-tune \emph{FLUX.1 Kontext}~\citep{labs2025flux} is motivated by its state-of-the-art FM-based architecture, which is a rectified flow transformer that unifies image generation and editing tasks with remarkable speed and consistency, making it an ideal foundation for our work.

\paragraph{Enhancing Semantic Control with MLLM Encoders.}
A final crucial advancement lies in improving the model's comprehension of user instructions.  The standard CLIP text encoders used in early models often fail to parse complex spatial, relational, or compositional commands~\citep{radford2021learning}.  To overcome this, a clear trend has emerged towards replacing or augmenting these encoders with powerful multimodal large language models (MLLMs).  State-of-the-art open-source models like \emph{Step1X-Edit}~\citep{liu2025step1x} and \emph{UniWorld-V1}~\citep{lin2025uniworld} have converged on a similar, highly effective architecture: both employ a potent MLLM, such as Qwen-VL~\citep{bai2025qwen2} or LLaVA~\citep{liu2023visual}, to process the input image and textual instruction, extracting a rich semantic embedding that conditions a DiT or FLUX-based generative backbone.  This approach leverages the MLLM's superior cross-modal reasoning to achieve more precise semantic control.  Our work adopts this SOTA paradigm by integrating Qwen-VL as a text and image encoder, demonstrating that the combination of an advanced flow matching architecture, enhanced semantic understanding via an MLLM, and a large-scale, high-quality refined dataset leads to significant performance gains.
\paragraph{Evaluation Benchmarks.}
We evaluate on four complementary instruction‑editing benchmarks: \emph{GEdit‑Bench‑EN (Full)}, \emph{ImgEdit (Full)}, \emph{Complex‑Edit}, and \emph{OmniContext}. GEdit‑Bench‑EN covers 11 edit types (background, color, material/texture, motion, portrait, style, add/remove/replace subject, text, tone) with MLLM‑based scoring of instruction following and perceptual quality~\citep{liu2025step1x}; ImgEdit spans 9 task families (Add, Adjust, Extract, Replace, Remove, Background, Style, Hybrid, Action) with a unified evaluation pipeline~\citep{ye2025imgedit}; Complex‑Edit composes chains of atomic edits to probe compositional reasoning and reports Instruction Following (IF), Identity Preservation (IP), and Perceptual Quality (PQ)~\citep{yang2025texttt}; OmniContext targets context‑aware in‑context generation/editing (single/multiple/scene) and uses GPT‑4.1 for interpretable, metric‑driven assessment~\citep{wu2025omnigen2}.

\section{Data Curation}
\begin{figure}[t!]
\begin{center}
\includegraphics[width=.72\linewidth]{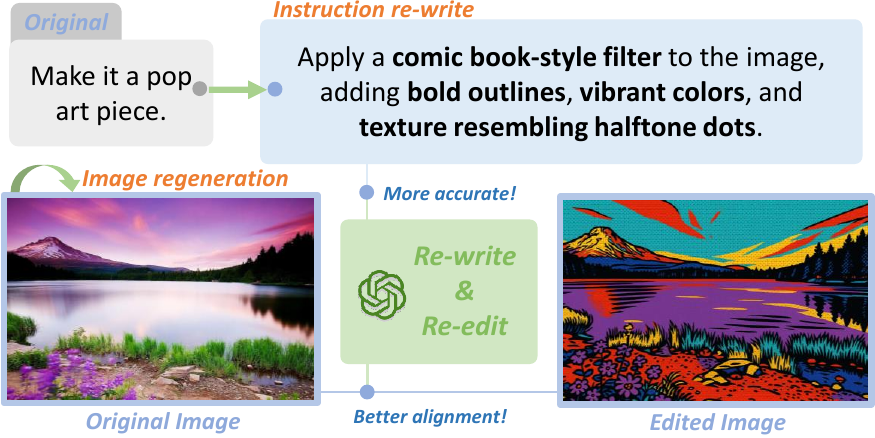}
\end{center}
\vspace{-1em}
\caption{An overview of \ourdata~data curation pipeline. We applied mutiple methods to collect high-quality image-editing data. We used GPT-4o to re-write 10\% instructions of the original OmniEdit dataset to make them more accurate, and the input images originally generated by DALL-E in HQ-Edit were re-synthesized by GPT-Image-1 for higher alignment.}
\vspace{-1em}
\label{fig:pipeline}
\end{figure}

As the core motivation of this work is to explore the plausibility of replacing currently widely-used data collection methods for instruction-based image editing, which depend on predefined editing operation types and manually crafted expert pipelines, with a minimalist pipeline that utilizes a pre-existing image editing model for various editing operations.
The data curation pipeline is shown in Fig~\ref{fig:pipeline}.

\subsection{Generation and Alignment Pipeline}
To create the final \ourdata~corpus, we merged OmniEdit, HQ-Edit, and UltraEdit, unifying their metadata into a consistent format. The core of our data generation process relied on the \texttt{gpt-image-1} API, which we used to regenerate or augment all image-instruction pairs. To maintain consistency, all generated images were created in high-quality mode and constrained to one of three fixed aspect ratios: 1:1 (1024$\times$1024), 3:2 (1536$\times$1024), or 2:3 (1024$\times$1536).

A key challenge was adapting source images with varying aspect ratios without introducing distortion or losing content. Naïve resizing would stretch objects, while simple cropping could remove semantically relevant regions. Our solution was a simple \emph{pad-and-crop} procedure: we snap each source image's aspect ratio to the nearest supported ratio and add minimal white padding to match the target dimensions. After generation, this padding is precisely cropped away. Finally, we apply automatic quality filtering, rejecting any outputs with artifacts or residual padding. Further dataset-specific details are provided in the Appendix.

\subsection{Dataset Specific Processes}
\paragraph{Instruction Re-writing}
For re-edited OmniEdit,  we used GPT-4o to rewrite part of the instructions with the original input image and the updated output image in~\ourdata. Under our cost limitation, we managed to rewrite $\sim10\%$ instructions of the original OmniEdit dataset.

\paragraph{Input Image Regeneration}
Since the input image in HQ-Edit is originally generated by DALL-E 3, which is an arguably outdated image generation model, we regenerated about $\sim50\%$ of the input images with GPT-Image-1 and produced corresponding output images based on these re-generated input images.

\paragraph{Complex-Edit style instruction}
One key benefit of a data collection pipeline that utilizes a general-purpose image-editing model instead of a set of expert pipelines, each for one of the manually crafted editing operations, is that it can easily handle complex, composed instructions. Under this motivation, we generate Complex-Edit style instructions with $\sim50\%$ input images of OmniEdit. As we noticed that edited images by GPT-Image-1 with very complex instructions tend to lose the realistic feel, we opted for $C3$ level of complexity, that is each complex instruction is composed with 3 atomic instructions.

\begin{table}[t!]
\centering
\caption{Comparison on the GEdit-EN-full benchmark. Our model achieves the highest average score among open-source methods.}
\label{tab:gedit}
\resizebox{\textwidth}{!}{%
\begin{tabular}{lcccccccccccc}
\toprule
\textbf{Model} &
\makecell{\textbf{BG}\\\textbf{Change}} &
\makecell{\textbf{Color}\\\textbf{Alt.}} &
\makecell{\textbf{Mat.}\\\textbf{Mod.}} &
\textbf{Motion} & \textbf{Portrait} & \textbf{Style} &
\textbf{Add} & \textbf{Remove} & \textbf{Replace} &
\textbf{Text} & \textbf{Tone} & \textbf{Avg} \\
\midrule
\multicolumn{3}{l}{\textit{Open-Sourced Models}} \\
AnyEdit & 4.31 & 4.25 & 2.64 & 0.67 & 1.90 & 1.95 & 3.72 & 3.75 & 3.23 & 0.77 & 4.21 & 2.85 \\
MagicBrush & 6.17 & 5.41 & 4.75 & 1.55 & 2.90 & 4.10 & 5.53 & 4.13 & 5.10 & 1.33 & 5.07 & 4.19 \\
Instruct-Pix2Pix & 3.94 & 5.40 & 3.52 & 1.27 & 2.62 & 4.39 & 3.07 & 1.50 & 3.48 & 1.13 & 5.10 & 3.22 \\
OmniGen & 5.23 & 5.93 & 5.44 & 3.12 & 3.17 & 4.88 & 6.33 & 6.35 & 5.34 & 4.31 & 4.96 & 5.01 \\
Step1X-Edit & 7.03 & 6.26 & 6.46 & 3.66 & 5.23 & 7.24 & 7.17 & 6.42 & 7.39 & 7.40 & 6.62 & 6.44 \\
Bagel & 7.44 & 6.99 & 6.26 & 5.09 & 4.82 & 6.04 & 7.94 & 7.37 & 7.31 & 7.16 & 6.17 & 6.60 \\
Bagel-thinking & 7.22 & 7.24 & 6.69 & 7.12 & 6.03 & 6.17 & 7.93 & 7.44 & 7.45 & 3.61 & 6.36 & 6.66 \\
Ovis-U1 & 7.49 & 6.88 & 6.21 & 4.79 & 5.98 & 6.46 & 7.49 & 7.25 & 7.27 & 4.48 & 6.31 & 6.42 \\
OmniGen2 & - & - & - & - & - & - & - & - & - & - & - & 6.42 \\
Step1X-Edit(v1.1) & 7.45 & 7.38 & 6.95 & 4.73 & 4.70 & 7.11 & 8.20 & 7.59 & 7.80 & 7.91 & 6.85 & 6.97 \\
FluxKontext dev & 7.06 & 7.03 & 5.52 & 5.62 & 4.68 & 5.55 & 6.95 & 6.76 & 6.13 & 6.10 & 7.48 & 6.26 \\
\midrule
\multicolumn{3}{l}{\textit{Proprietary Models}} \\
Gemini & 7.11 & 7.14 & 6.47 & 5.67 & 3.99 & 4.95 & 8.12 & 6.89 & 7.41 & 6.85 & 7.01 & 6.51 \\
Doubao & 8.07 & 7.36 & 7.20 & 5.38 & 6.28 & 7.20 & 8.05 & 7.71 & 7.87 & 4.01 & 7.67 & 6.98 \\
GPT-4o & 6.96 & 6.85 & 7.10 & 5.41 & 6.74 & 7.44 & 7.51 & 8.73 & 8.55 & 8.45 & 8.69 & 7.49 \\
\midrule
\rowcolor{cyan!10} \textbf{Ours} & {7.80} & {7.54} & {7.12} & {7.75} & {7.09} & {6.74} & {8.04} & {7.95} & {7.17} & {5.45} & {6.95} & {7.24} \\
\bottomrule
\end{tabular}%
}
\end{table}

\begin{table}[t!]
\centering
\caption{Comparison on the ImgEdit-Full benchmark.}
\label{tab:imgedit}
\resizebox{\textwidth}{!}{%
\begin{tabular}{lcccccccccc}
\toprule
\textbf{Model} & \textbf{Add} & \textbf{Adjust} & \textbf{Extract} & \textbf{Replace} & \textbf{Remove} & \textbf{Background} & \textbf{Style} & \textbf{Hybrid} & \textbf{Action} & \textbf{Overall} \\
\midrule
MagicBrush & 2.84 & 1.58 & 1.51 & 1.97 & 1.58 & 1.75 & 2.38 & 1.62 & 1.22 & 1.90 \\
Instruct-Pix2Pix & 2.45 & 1.83 & 1.44 & 2.01 & 1.50 & 1.44 & 3.55 & 1.20 & 1.46 & 1.88 \\
AnyEdit & 3.18 & 2.95 & 1.88 & 2.47 & 2.23 & 2.24 & 2.85 & 1.56 & 2.65 & 2.45 \\
UltraEdit & 3.44 & 2.81 & 2.13 & 2.96 & 1.45 & 2.83 & 3.76 & 1.91 & 2.98 & 2.70 \\
OmniGen & 3.47 & 3.04 & 1.71 & 2.94 & 2.43 & 3.21 & 4.19 & 2.24 & 3.38 & 2.96 \\
Step1X-Edit & 3.88 & 3.14 & 1.76 & 3.40 & 2.41 & 3.16 & 4.63 & 2.64 & 2.52 & 3.06 \\
ICEdit & 3.58 & 3.39 & 1.73 & 3.15 & 2.93 & 3.08 & 3.84 & 2.04 & 3.68 & 3.05 \\
BAGEL & 3.56 & 3.31 & 1.70 & 3.30 & 2.62 & 3.24 & 4.49 & 2.38 & 4.17 & 3.20 \\
UniWorld-V1 & 3.82 & 3.64 & 2.27 & 3.47 & 3.24 & 2.99 & 4.21 & 2.96 & 2.74 & 3.26 \\
OmniGen2 & 3.57 & 3.06 & 1.77 & 3.74 & 3.20 & 3.57 & 4.81 & 2.52 & 4.68 & 3.44 \\
Ovis-U1 & 4.13 & 3.62 & 2.98 & 4.45 & 4.06 & 4.22 & 4.69 & 3.45 & 4.61 & 4.00 \\
FluxKontext dev & 3.76 & 3.45 & 2.15 & 3.98 & 2.94 & 3.78 & 4.38 & 2.96 & 4.26 & 3.52 \\
\midrule
GPT-4o & 4.61 & 4.33 & 2.90 & 4.35 & 3.66 & 4.57 & 4.93 & 3.96 & 4.89 & 4.20 \\
\midrule
\rowcolor{cyan!10} \textbf{Ours} & {4.07} & {3.79} & {2.04} & {4.13} & {3.89} & {3.90} & {4.84} & {3.04} & {4.52} & {3.80} \\
\bottomrule
\end{tabular}%
}
\end{table}

\section{Experiments}

\subsection{Experimental Setup}
\paragraph{Models}
Our primary model, which we refer to as \emph{Ours}, is built upon the state-of-the-art \emph{FluxKontext dev} framework. We enhance its performance by replacing the original CLIP-based encoders with \emph{Qwen-VL-7b} embeddings for both image and instruction conditioning, which improves semantic alignment. For our ablation studies, we also evaluate a \emph{SD3-Medium} model, which uses a DiT-style architecture with channel-wise conditioning, and the original \emph{Flux 1.0 dev} model, which uses token-wise control with SigLIP features \citep{zhai2023sigmoid}.

\paragraph{Benchmarks}
We evaluate our models on a comprehensive suite of benchmarks to assess various editing capabilities. These include \emph{GEdit-EN-full} and \emph{ImgEdit-Full} for general editing performance, \emph{Complex-Edit} for compositional understanding, and \emph{OmniContext} for context-aware editing.

\subsection{Main Results}
As demonstrated in Tables \ref{tab:gedit}, \ref{tab:imgedit}, \ref{tab:complex} and \ref{tab:omni_single}, our model, trained on the \ourdata~dataset, achieves state-of-the-art performance among open-source methods and is highly competitive with leading proprietary models like GPT-4o. On GEdit-EN-full, our model scores an average of \textbf{7.236}, outperforming all other open-source models. Similarly, on ImgEdit-Full, our model achieves an overall score of \textbf{3.80}, again surpassing existing methods. 
In the challenging Complex-Edit benchmark with $C8$ complexity, which decomposes evaluation into Instruction Following (IF), Identity Preservation (IP), and Perceptual Quality (PQ), our model scores an impressive \textbf{8.78} overall, demonstrating a strong balance between accurately following instructions (8.99 IF) and preserving unchanged content (8.41 IP).

\begin{table}[t!]
\centering
\caption{Comparison on the Complex-Edit benchmark.}
\label{tab:complex}
\resizebox{.45\textwidth}{!}{%
\begin{tabular}{lcccc}
\toprule
\textbf{Method} & \textbf{IF} & \textbf{IP} & \textbf{PQ} & \textbf{Overall} \\
\midrule
AnyEdit & 1.60 & 8.15 & 7.25 & 5.67 \\
UltraEdit & 6.56 & 5.93 & 7.29 & 6.59 \\
OmniGen & 6.25 & 6.42 & 7.54 & 6.74 \\
FluxKontext Dev & 8.56 & 8.39 & 8.51 & 8.49 \\
\midrule
Imagen3 & 7.56 & 6.55 & 7.67 & 7.26 \\
SeedEdit & 8.49 & 6.91 & 8.74 & 8.04 \\
GPT-4o & 9.29 & 7.51 & 9.47 & 8.76 \\
\midrule
\rowcolor{cyan!10} \textbf{Ours} & {8.99} & {8.41} & {8.93} & {8.78} \\
\bottomrule
\end{tabular}
}
\end{table}

\begin{table}[t!]
\centering
\caption{Results on the OmniContext \textit{SINGLE} benchmark.}
\label{tab:omni_single}
\resizebox{.85\textwidth}{!}{%
\begin{tabular}{lccc ccc ccc}
\toprule
\multirow{2}{*}{\textbf{Method}} & \multicolumn{3}{c}{\textbf{Character}} & \multicolumn{3}{c}{\textbf{Object}} & \multicolumn{3}{c}{\textbf{Average}} \\
\cmidrule(lr){2-4}\cmidrule(lr){5-7}\cmidrule(lr){8-10}
& \textbf{PF} & \textbf{SC} & \textbf{Overall} & \textbf{PF} & \textbf{SC} & \textbf{Overall} & \textbf{PF} & \textbf{SC} & \textbf{Overall} \\
\midrule
InfiniteYou         & 7.81 & 5.15 & 6.05 & \textemdash & \textemdash & \textemdash & \textemdash & \textemdash & \textemdash \\
UNO                 & 7.56 & 6.48 & 6.60 & 7.78 & 6.65 & 6.83 & 7.67 & 6.56 & 6.72 \\
BAGEL               & 7.72 & 4.86 & 5.48 & 8.56 & 6.06 & 7.03 & 8.14 & 5.46 & 6.25 \\
OmniGen             & 7.12 & 7.58 & 7.21 & 7.66 & 5.04 & 5.71 & 7.39 & 6.31 & 6.46 \\
OmniGen2            & 8.04 & 8.34 & 8.05 & 8.44 & 7.26 & 7.58 & 8.24 & 7.80 & 7.81 \\
Flux.1 Kontext (dev)& 7.70 & 8.72 & 8.07 & 8.76 & 8.22 & 8.33 & 8.23 & 8.47 & 8.20 \\
\midrule
Flux.1 Kontext (max) & 7.98 & 9.24 & 8.48 & 8.78 & 8.76 & 8.68 & 8.38 & 9.00 & 8.58 \\
Gemini-2.0-Flash    & 5.54 & 5.98 & 5.06 & 6.17 & 5.89 & 5.17 & 5.86 & 5.93 & 5.11 \\
GPT-4o              & 8.89 & 9.03 & 8.90 & 9.40 & 8.74 & 9.01 & 9.14 & 8.88 & 8.95 \\
\midrule
\rowcolor{cyan!10}\textbf{Ours} & 8.10 & 8.36 & 8.11 & 8.50 & 7.68 & 7.87 & 8.30 & 8.02 & 7.99 \\
\bottomrule
\end{tabular}%
}
\end{table}

\subsection{Ablation Studies}
We conduct a series of ablation studies to dissect the sources of performance improvement, isolating the effects of our data curation strategies and our model architecture choices.

\paragraph{Impact of Data Curation Strategies}
Table \ref{tab:data_ablation} shows the impact of our different data refinement strategies. Across all experiments, training on the baseline datasets yields the lowest scores. Our first refinement step, regenerating only the output image (\texttt{-gpt} or \texttt{-output-regen}), provides a substantial performance uplift. For instance, with Flux 1.0 dev on OmniEdit, this step improves the GEdit-EN score from 4.93 to 5.98. Our second step, regenerating the instruction to match the new output (\texttt{-gpt-rewrite}), provides a further boost to instruction-following capabilities, raising the imgedit score from 3.24 to 3.40. This validates our meticulous, multi-step approach to data curation.

\begin{table}[h!]
\centering
\caption{Ablation studies on data curation strategies. We report imgedit and GEdit-EN scores for models trained on different 100k data variants.}
\label{tab:data_ablation}
\resizebox{.65\textwidth}{!}{%
\begin{tabular}{llcc}
\toprule
\textbf{Method} & \textbf{Dataset Variant} & \textbf{imgedit} & \textbf{GEdit-EN} \\
\midrule
\multicolumn{4}{c}{\textit{OmniEdit Ablations}} \\
\midrule
SD3-Medium & omniedit100k-base & 2.54 & 3.92 \\
SD3-Medium & omniedit100k-gpt & 3.13 & 4.91 \\
SD3-Medium & omniedit100k-gpt-rewrite & 3.32 & 4.89 \\
\midrule
Flux 1.0 dev & omniedit100k-base & 2.94 & 4.93 \\
Flux 1.0 dev & omniedit100k-gpt & 3.24 & 5.98 \\
Flux 1.0 dev & omniedit100k-gpt-rewrite & {3.40} & {5.88} \\
\midrule
\multicolumn{4}{c}{\textit{HQ-Edit Ablations}} \\
\midrule
SD3-Medium & hqedit100k-base & 2.19 & 2.00 \\
SD3-Medium & hqedit100k-output-regen & 3.02 & 4.45 \\
SD3-Medium & hqedit100k-pair-regen & {3.08} & {4.75} \\
\midrule
Flux 1.0 dev & hqedit100k-base & 3.12 & 4.34 \\
Flux 1.0 dev & hqedit100k-output-regen & 3.44 & 5.67 \\
Flux 1.0 dev & hqedit100k-pair-regen & {3.45} & {5.73} \\
\midrule
\multicolumn{4}{c}{\textit{Complex-Edit Instruction Ablation}} \\
\midrule
Flux 1.0 dev & Complex-Edit & 2.89 & 5.39 \\
\bottomrule
\end{tabular}
}
\end{table}

Across all experiments, training on the baseline datasets yields the lowest scores. Our first refinement step, regenerating only the output image (\texttt{-gpt} or \texttt{-output-regen}), provides a substantial performance uplift. For instance, with Flux 1.0 dev on OmniEdit, this step improves the GEdit-EN score from 4.93 to 5.98. Our second step, regenerating the instruction to match the new output (\texttt{-gpt-rewrite}), provides a further boost to instruction-following capabilities, raising the imgedit score from 3.24 to 3.40.

\begin{table}[t!]
\centering
\caption{Ablation on the inclusion of the Complex-Edit data subset on the GEdit-EN-full benchmark.}
\label{tab:complex_ablation}
\resizebox{\textwidth}{!}{%
\begin{tabular}{lcccccccccccc}
\toprule
\textbf{Dataset} &
\makecell{\textbf{BG}\\\textbf{Change}} &
\makecell{\textbf{Color}\\\textbf{Alt.}} &
\makecell{\textbf{Mat.}\\\textbf{Mod.}} &
\textbf{Motion} & \textbf{Portrait} & \textbf{Style} &
\textbf{Add} & \textbf{Remove} & \textbf{Replace} &
\textbf{Text} & \textbf{Tone} & \textbf{Avg} \\
\midrule
Ours w/o complex & 7.62 & 7.55 & 6.77 & 7.08 & 6.74 & 6.74 & 7.68 & 7.74 & 6.82 & 5.36 & 7.23 & 7.03 \\
Ours (full) & {7.80} & {7.54} & {7.12} & {7.75} & {7.09} & {6.74} & {8.04} & {7.95} & {7.17} & {5.45} & {6.95} & {7.24} \\
\bottomrule
\end{tabular}%
}
\end{table}

\begin{table}[t!]
\centering
\caption{Ablation on the inclusion of the Complex-Edit data subset on the ImgEdit benchmark.}
\label{tab:complex_ablation_imgedit}
\resizebox{\textwidth}{!}{%
\begin{tabular}{lcccccccccc}
\toprule
\textbf{Dataset} & \textbf{Add} & \textbf{Adjust} & \textbf{Extract} & \textbf{Replace} & \textbf{Remove} & \textbf{Background} & \textbf{Style} & \textbf{Hybrid} & \textbf{Action} & \textbf{Overall} \\
\midrule
Ours w/o complex & 4.07 & 3.69 & 1.94 & 4.17 & 3.93 & 3.73 & 4.74 & 2.91 & 4.19 & 3.71 \\
Ours (full)      & 4.07 & 3.79 & 2.04 & 4.13 & 3.89 & 3.90 & 4.84 & 3.04 & 4.52 & 3.80 \\
\bottomrule
\end{tabular}%
}
\end{table}

\begin{table}[t!]
\centering
\caption{Ablation on different text encoder configurations on the GEdit-EN-full benchmark.}
\label{tab:encoder_ablation}
\resizebox{\textwidth}{!}{%
\begin{tabular}{lcccccccccccc}
\toprule
\textbf{Text Encoder} &
\makecell{\textbf{BG}\\\textbf{Change}} &
\makecell{\textbf{Color}\\\textbf{Alt.}} &
\makecell{\textbf{Mat.}\\\textbf{Mod.}} &
\textbf{Motion} & \textbf{Portrait} & \textbf{Style} &
\textbf{Add} & \textbf{Remove} & \textbf{Replace} &
\textbf{Text} & \textbf{Tone} & \textbf{Avg} \\
\midrule
FluxKontext dev (T5) & 7.06 & 7.03 & 5.52 & 5.62 & 4.68 & 5.55 & 6.95 & 6.76 & 6.13 & 6.10 & 7.48 & 6.26 \\
Finetuned with T5 & 7.39 & 7.43 & 7.07 & 6.29 & 6.91 & 6.62 & 7.84 & 7.36 & 7.17 & 6.22 & 8.04 & 7.12 \\
Finetuned with QwenVL & 6.45 & 7.27 & 5.04 & 6.53 & 7.26 & 5.88 & 7.03 & 7.20 & 4.31 & 1.20 & 6.64 & 5.89 \\
QwenVL + T5 (Ours) & {7.80} & {7.54} & {7.12} & {7.75} & {7.09} & {6.74} & {8.04} & {7.95} & {7.17} & {5.45} & {6.95} & {7.24} \\
\bottomrule
\end{tabular}%
}
\end{table}

\begin{table}[t!]
\centering
\caption{Ablation on different text encoder configurations on the ImgEdit benchmark.}
\label{tab:encoder_ablation_imgedit}
\resizebox{\textwidth}{!}{%
\begin{tabular}{lccccccccccc}
\toprule
\textbf{Text Encoder} & \textbf{Add} & \textbf{Adjust} & \textbf{Extract} & \textbf{Replace} & \textbf{Remove} & \textbf{Background} & \textbf{Style} & \textbf{Hybrid} & \textbf{Action} & \textbf{Overall} \\
\midrule
FluxKontext dev (T5) & 3.76 & 3.45 & 2.15 & 3.98 & 2.94 & 3.78 & 4.38 & 2.96 & 4.26 & 3.52 \\
Finetuned with T5 & 4.19 & 3.79 & 2.09 & 4.22 & 3.96 & 3.90 & 4.76 & 3.23 & 4.49 & 3.85 \\
Finetuned with QwenVL & 3.92 & 3.58 & 1.95 & 3.62 & 3.89 & 3.72 & 4.64 & 3.22 & 3.82 & 3.60 \\
QwenVL + T5 (Ours) & {4.07} & {3.79} & {2.04} & {4.13} & {3.89} & {3.90} & {4.84} & {3.04} & {4.52} & {3.80} \\
\bottomrule
\end{tabular}%
}
\end{table}
Interestingly, our ablation on using raw complex instructions (\texttt{Complex-Edit}) resulted in relatively poor performance (5.39 on GEdit-EN). Visual inspection of the outputs revealed a significant failure in identity preservation, where the model would alter uninstructed parts of the image. This critical finding underscores that merely increasing instruction complexity is insufficient and can even be detrimental if not paired with high-quality, aligned image pairs. This validates our meticulous, multi-step approach to data curation, which explicitly optimizes for instruction following, identity preservation, and perceptual quality.

\paragraph{Impact of Complex-Edit Data and Text Encoders}
We further analyze the impact of including the Complex-Edit subset and the choice of text encoder. As shown in Table \ref{tab:complex_ablation} and \ref{tab:complex_ablation_imgedit}, including the Complex-Edit data provides a consistent, measurable boost across both GEdit-EN and ImgEdit benchmarks, improving the average scores from 7.03 to 7.24 and 3.71 to 3.80, respectively. This highlights the value of training on more challenging, compositional instructions.

Tables \ref{tab:encoder_ablation} and \ref{tab:encoder_ablation_imgedit} ablate the choice of instruction text encoder with all text encoders frozen (T5, Qwen-VL, and CLIP). We fine-tune the rest of FluxKontext end-to-end. Using frozen T5 already lifts GEdit-EN from 6.26 to 7.12. Using frozen Qwen-VL alone underperforms on the Text category 1.20, likely due to tokenizer merges that hinder isolating target strings. Concatenating frozen Qwen-VL and frozen T5 features yields the best GEdit-EN average 7.24 and competitive ImgEdit overall 3.80, while retaining CLIP’s pooled global features for architectural parity.

\begin{figure}[t!]
\begin{center}
\includegraphics[width=1\linewidth]{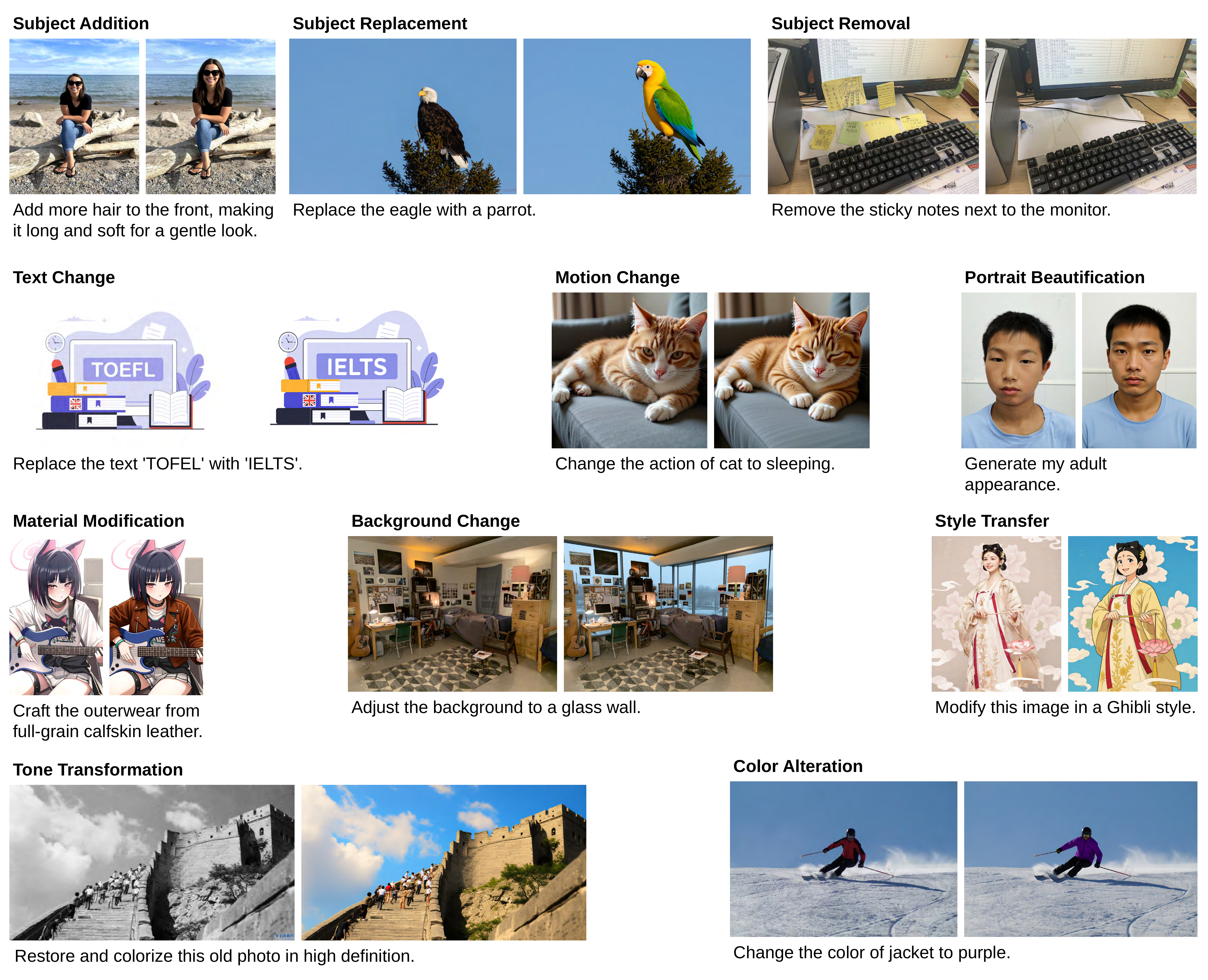}
\end{center}
\vspace{-1.5em}
\caption{The qualitative results of our method on G-Edit-Benchmark-EN.}
\label{fig:g-edit-benchmark-en}
\vspace{-1em}
\end{figure}

\begin{figure}[t!]
\begin{center}
\includegraphics[width=\linewidth]{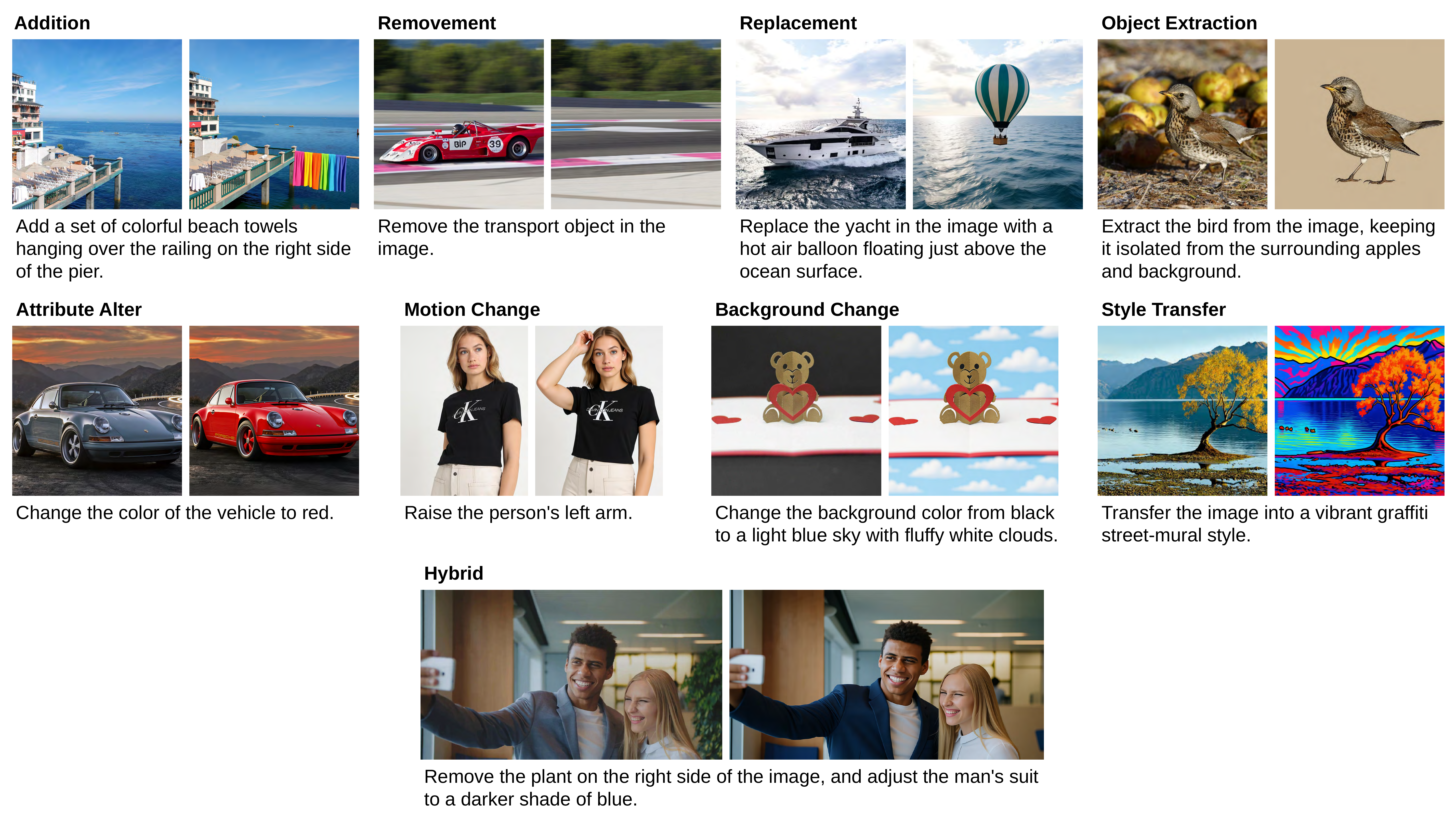}
\end{center}
\caption{The qualitative results of our method on Img-Edit.}
\label{fig:img-edit}
\end{figure}

\begin{figure}[t!]
\begin{center}
\includegraphics[width=\linewidth]{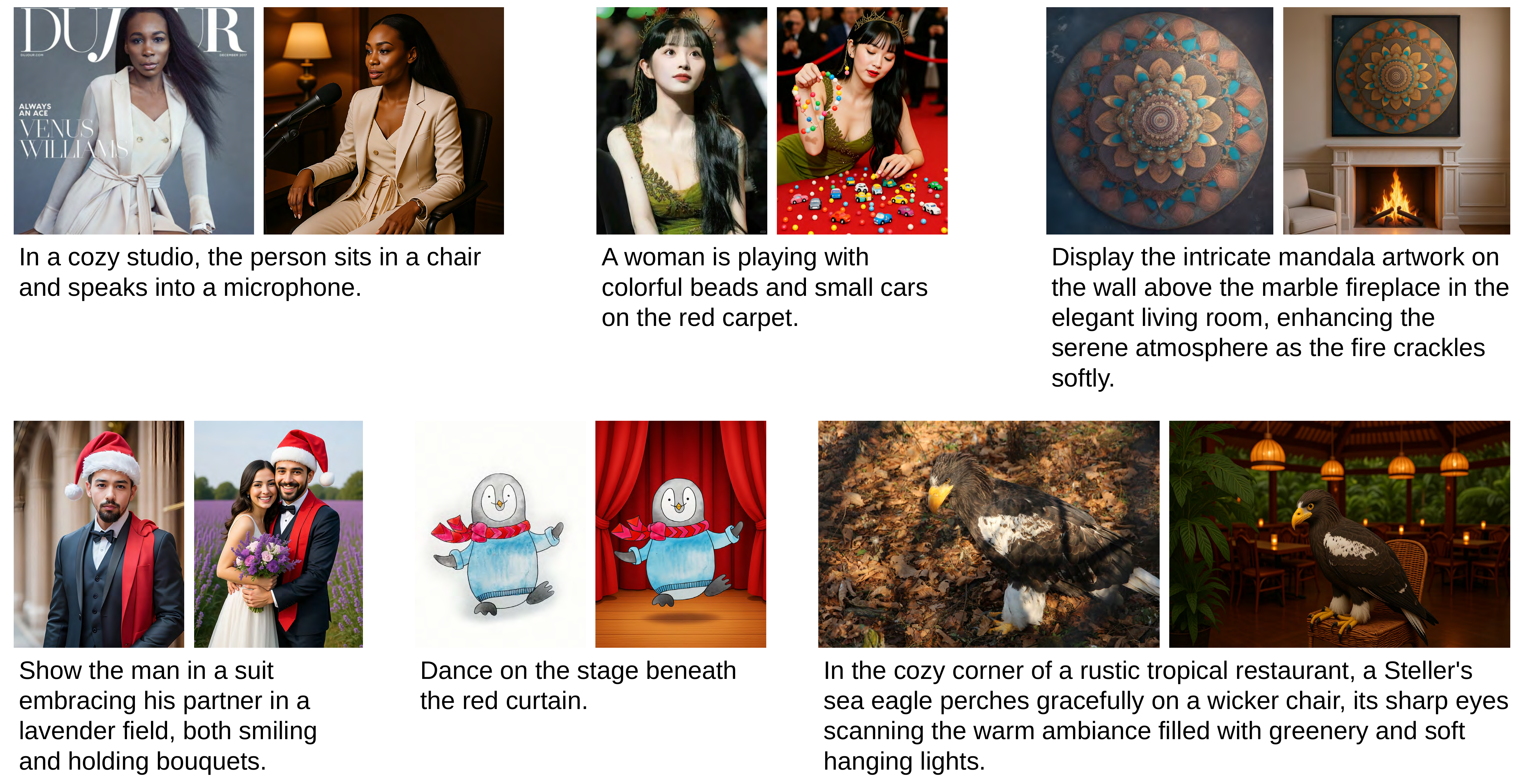}
\end{center}
\vspace{-1em}
\caption{The qualitative results of our method on OmniContext.}
\label{fig:omnicontext}
\end{figure}

\begin{figure}[t!]
\begin{center}
\includegraphics[width=\linewidth]{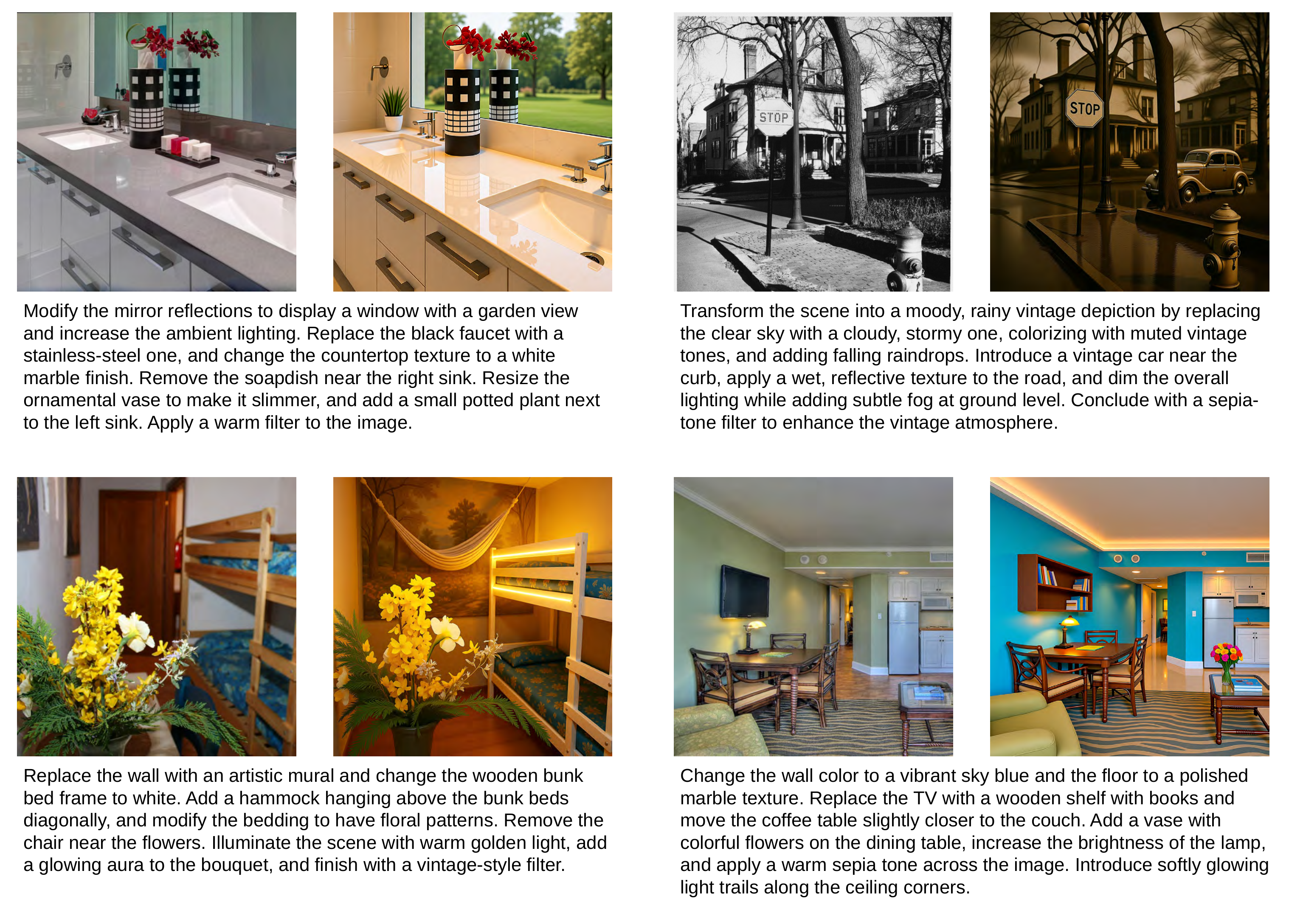}
\end{center}
\vspace{-1em}
\caption{The qualitative results of our method on Complex-Edit.}
\label{fig:complex-edit}
\end{figure}

\subsection{Qualitative Results}
In Fig~\ref{fig:g-edit-benchmark-en}, \ref{fig:img-edit}, \ref{fig:omnicontext}, and \ref{fig:complex-edit}, we present the qualitative editing results of the model trained on \ourdata on different types editing scenarios as defined by these benchmarks.
It is clear that the model trained on \ourdata demonstrates a good understanding of the editing instruction as well as generating realistic images while keeping elements that are not instructed to change the same.

\section{Conclusion}
In this work, we addressed the critical need for a large-scale, high-quality dataset to advance open-source instruction-based image editing. We introduced \ourdata, a corpus of over 1.5 million samples created by systematically refining and unifying existing datasets using GPT-4o. Our data curation process was explicitly guided by the core principles of a successful edit: instruction following, identity preservation, and perceptual quality.
Our experiments demonstrate the significant impact of this new dataset. By fine-tuning a state-of-the-art open-source model on \ourdata, we achieved new SOTA performance across multiple benchmarks, substantially narrowing the gap with proprietary models. The ablation studies validated our multi-faceted data generation strategy, showing that each refinement step—from output regeneration to instruction rewriting—provides tangible benefits. Crucially, we also highlighted that data quality, specifically the tight alignment between instruction and image while preserving identity, is more important than mere instruction complexity.
By releasing the \ourdata~dataset and our fine-tuned models, we provide a powerful resource for the research community. We hope this work will catalyze further innovation in open-source image editing, enabling the development of models with even greater capabilities. Future work could explore applying this data curation methodology to other modalities like video or 3D, or investigate more automated techniques for detecting and correcting subtle misalignments in generative data pipelines.

\section{Acknowledge}
We would like to thank 	Ashwin Nagarajan, Tejas Polu, Jiawei Mao, Zeyu Wang and Haoqin Tu for the early discussion and exploration of this project. We would like to thank the Microsoft Accelerate Foundation Models Research Program for supporting our computing needs.

\bibliography{iclr2025_conference}
\bibliographystyle{iclr2025_conference}

\clearpage
\appendix
\section{Dataset-Specific Processing Details}

\subsection{UltraEdit Downscale Workflow}
The UltraEdit dataset originally uses 512$\times$512 inputs. Our workflow was to regenerate both the input (where applicable) and the output at 1024$\times$1024 using our standard procedure, and then downsample both images back to 512$\times$512 using bicubic interpolation to maintain compatibility with the original benchmark's expectations.

\subsection{OmniEdit Alignment Procedure}
For each sample in OmniEdit, we applied our standard geometric alignment: compute the image's ratio, pad to the nearest supported generation ratio, and run the edit. After generation, we crop the padding and resize the image back to its original dimensions to ensure comparable pixel density. We implemented a strict quality filter, rejecting any sample if more than 0.5\% of its border consisted of uniform padding after the process.

\subsection{Complex-Edit Subset}
The Complex-Edit subset was handled similarly to OmniEdit regarding the geometry and padding procedure. It contains only the canonical complex instructions. We applied a stringent filter, discarding any output that had a detectable padding error after the crop step.

\subsection{HQ-Edit Dual Splits}
The HQ-Edit portion of our dataset was processed in two distinct splits:
\begin{description}
    \item[Edit Split] For existing pairs, we pad the original input image, generate the edit, crop the padding, and restore the image to its original resolution.
    \item[Generate Split] For generation tasks, we first synthesize a new reference input image from the textual instruction. We then apply the same edit instruction to this newly generated input. For these tasks, the aspect ratio was chosen randomly from our three supported options (1:1, 2:3, 3:2) to increase diversity.
\end{description}

\end{document}

%% file: math_commands.tex
\usepackage{amsmath,amsfonts,bm}

\def\eqref#1{equation~\ref{#1}}

\def\1{\bm{1}}

\DeclareMathAlphabet{\mathsfit}{\encodingdefault}{\sfdefault}{m}{sl}
\SetMathAlphabet{\mathsfit}{bold}{\encodingdefault}{\sfdefault}{bx}{n}